\documentclass[letterpaper]{article}
\usepackage{aaai2027}
\usepackage[hyphens]{url}
\usepackage{graphicx}
\usepackage{natbib}
\usepackage{caption}
\usepackage{amsmath}
\usepackage{amssymb}
\usepackage{booktabs}
\usepackage{algorithm}
\usepackage{algpseudocode}
\usepackage{graphicx}
\usepackage{natbib}
\usepackage{amsmath}
\usepackage{amssymb}
\usepackage{booktabs}
\usepackage{caption}

\title{The Time Value of Evolution}

\author{
\begin{minipage}[t]{0.30\textwidth}
    \centering
    \textbf{Matthew Siper}\\[3pt]
    \small
    Nof1, New York University\\
    \texttt{siper.matthew@gmail.com}
\end{minipage}
\hfill
\begin{minipage}[t]{0.30\textwidth}
    \centering
    \textbf{Ahmed Khalifa}\\[3pt]
    \small
    Nof1, University of Malta\\
    \texttt{ahmed.khalifa@um.edu.mt}
\end{minipage}
\hfill
\begin{minipage}[t]{0.30\textwidth}
    \centering
    \textbf{Julian Togelius}\\[3pt]
    \small
    Nof1, New York University\\
    \texttt{julian.togelius@gmail.com}
\end{minipage}
}

\affiliations{}

\begin{document}
\maketitle

\begin{abstract}
In evolutionary search, a weak child can be a valuable ancestor that makes high-fitness regions reachable. Immediate-return control is blind to this delayed utility, penalizing mutations through their immediate offspring even when they open productive future lineages. We formalize this hidden dynamic as the \emph{time value of evolution} within a finite-horizon Markov decision process. To exploit it, we introduce \emph{Lineage-Value Policy Gradients} (LVPG), a long-horizon actor-critic framework for automated trading policy discovery. Our architecture decouples search control into specialized policy heads over a shared generative backbone: a bootstrapped critic head estimates the value of finite-horizon lineage potential from multi-step mutation trees, while an actor head dynamically modulates mutation intensity over the remaining search budget. We isolate the impact of long-horizon credit assignment against immediate-return optimization across 90 paired runs under matched operators, lineage supervision, folds, seeds, and budgets. Path-based credit assignment substantially accelerates finite-budget search, increasing validation best-so-far AUC by $0.394$ Sharpe units. LVPG also produces fewer temporary regressions than immediate-return optimization and recovers from them more often. Finite-horizon lineage value yields more selective non-monotonic search and stronger policies within identical resource constraints.
\end{abstract}

\section{Introduction}
\label{sec:introduction}
Evolutionary search often treats the fitness of a newly generated child as the return for the mutation that produced it. This is adequate when useful edits improve fitness immediately, but it undervalues structural changes that first lower fitness and later enable stronger descendants. Semantic program mutation makes this mismatch especially important because one edit can alter indicators, control flow, and decision logic. Its value may also depend on the remaining budget. A wider mutation early in an eight-step lineage can be repaired and extended, while the same move near the horizon may expire before its downstream value appears.

\subsection{The Time Value of Evolution}

Option pricing separates current exercise value from the value preserved by time to expiry. Evolutionary mutations have an analogous property. A weak offspring can retain value by opening a productive region that later descendants can exploit before the search budget expires.

We formalize this delayed utility as the \emph{time value of evolution}. Let \(h=H-t\) denote the number of search steps remaining, and let
\begin{equation}
Q_h^{\pi}(s,z)
=
\mathbb{E}_{\pi}
\left[
B_{t+h}-B_t
\mid
s_t=s,\,
z_t=z
\right]
\end{equation}
denote the expected best-so-far improvement obtained by selecting mutation intensity \(z\) in state \(s\) and continuing under policy \(\pi\) for the remaining horizon. Its one-step counterpart is \(Q_1^{\pi}(s,z)\). We define
\begin{equation}
\operatorname{TVE}_h(s,z)
=
Q_h^{\pi}(s,z)
-
Q_1^{\pi}(s,z).
\label{eq:tve}
\end{equation}
A positive value means that one-step evaluation understates a mutation's finite-budget utility. Our primary objective uses \(\gamma=1\), so improvements are not discounted solely because they appear several mutations after the enabling action. The control problem is therefore to select the behavioral displacement with the greatest continuation value for the current program and remaining budget.

\subsection{Lineage-Value Policy Gradients}

We introduce \emph{Lineage-Value Policy Gradients} (LVPG), a long-horizon actor-critic framework for automated policy discovery. A frozen, offline-trained Evolutionary Language Model (ELM) generates and compiles mutations at three behaviorally calibrated radii, \textsc{Refine}, \textsc{Interpolate}, and \textsc{Explore}. Separate actor and critic LoRA adapters control the radius and estimate finite-horizon lineage value from the current program, evaluation state, and remaining budget.

The primary comparison isolates the temporal scope of policy-gradient credit. PPO-Path and PPO-Immediate share the ELM, state and action spaces, critic initialization, mutation-tree labels, auxiliary first-action supervision, folds, seeds, optimization settings, and evaluation budgets. They differ only in the return supplied to policy optimization. PPO-Path uses downstream best-so-far progress, while PPO-Immediate uses clipped one-step fitness change with \(\gamma=0\).

\paragraph{Contributions.}
\begin{itemize}
    \setlength{\itemsep}{0pt}
    \setlength{\parskip}{0pt}
    \setlength{\parsep}{0pt}

    \item \textbf{Conceptual.}
    We formalize the time value of evolution, distinguishing immediate offspring value from the finite-horizon value of reachable lineages.

    \item \textbf{Methodological.}
    We introduce LVPG, which combines state- and budget-dependent mutation control, a mutation-tree-bootstrapped critic, and policy-gradient credit over realized evolutionary paths.

    \item \textbf{Empirical.}
    Across a matched paired protocol, long-horizon credit improves search efficiency and sealed-test performance while producing fewer and more recoverable temporary regressions.
\end{itemize}

\section{Related Work}
\subsection{Adaptive Operator Control}
Parameter control and adaptive operator selection study how evolutionary algorithms should choose mutation settings online \citep{eiben1999parameter,karafotias2015parameter,fialho2010bandit}. Self-adaptive step sizes encode control variables within the evolving population, while bandit and reinforcement-learning methods update operator preferences from observed rewards. Delayed rewards, finite-horizon control, and policy gradients are therefore not new in isolation. Our claim is narrower. Most practical credit signals remain dominated by recent offspring improvement, and prior operator-control results do not establish whether a language-model-generated program mutation should be valued through the executable lineage it opens. We hold lineage supervision fixed and vary only the return used by PPO.

Novelty Search and MAP-Elites preserve stepping stones through behavioral diversity or structured archives \citep{lehman2011novelty,mouret2015mapelites}. Our method leaves the task objective, parent sampling, and archive rule unchanged. A lower-fitness child receives useful credit only when later descendants exceed the previous path best.

\subsection{Language-Model Variation}
Language models now serve as mutation, crossover, and refinement operators in evaluator-guided search. EvoPrompting applies them to neural architecture evolution, while FunSearch, Evolution of Heuristics, LLaMEA, and AlphaEvolve improve executable programs or algorithms through repeated evaluation \citep{chen2023evoprompting,romeraparedes2024funsearch,liu2024eoh,vanstein2025llamea,novikov2025alphaevolve}. These systems establish the value of language models as semantic variation engines. Our ELM is trained offline as a multitask learner for mutation, compilation, and translation, then behaviorally calibrated with Offset Direct Preference Optimization and frozen during PPO. Online learning is confined to separate actor and critic LoRA adapters and their output heads.

\subsection{Trading Policy Search}
Trading-rule evolution provides a difficult executable search setting with nonlinear program behavior, regime variation, transaction costs, and substantial selection risk \citep{allen1999genetic,white2000reality}. ProFiT evolves source-level trading programs with language-model feedback, while Continuous Program Search learns a geometry for behaviorally local variation \citep{siper2026profit,siper2026continuous}. Those works address semantic competence and spatial locality. The present work addresses temporal credit. Rolling-origin folds, sealed test windows, and complete-run inference are used to separate search progress from financial generalization.

\section{Evolutionary MDP}
\subsection{Environment}
We define
\begin{equation}
 \mathcal{M}_{\mathrm{evo}}=(\mathcal{S},\mathcal{Z},P,R,H), \qquad H=8.
 \label{eq:mdp}
\end{equation}
At step $t$, the rendered state prompt represents
\begin{equation}
 s_t=\left(p_t^{\mathrm{NL}},p_t^{\mathrm{GPTL}},m_t,B_t,t/H\right),
 \quad B_t=\max_{0\leq j\leq t}F(p_j),
 \label{eq:state}
\end{equation}
where $m_t$ contains training-window execution metrics and feedback. The discrete action space is
\begin{equation}
 \mathcal{Z}=\{\textsc{Refine},\textsc{Interpolate},\textsc{Explore}\}.
 \label{eq:actions}
\end{equation}
The transition uses two ELM tasks and one fixed evaluator. The frozen ELM, denoted $M_{\theta}$, samples $p_{t+1}^{\mathrm{NL}}\sim M_{\theta}(\cdot\mid s_t,z_t)$ and compiles the result into typed GPTL; the backtester returns $F(p_{t+1})$ and the next feedback state. Generation, compilation, execution, and path-best updating together define $P(s_{t+1}\mid s_t,z_t)$.

\begin{figure*}[t]
    \centering
    \includegraphics[width=\textwidth]{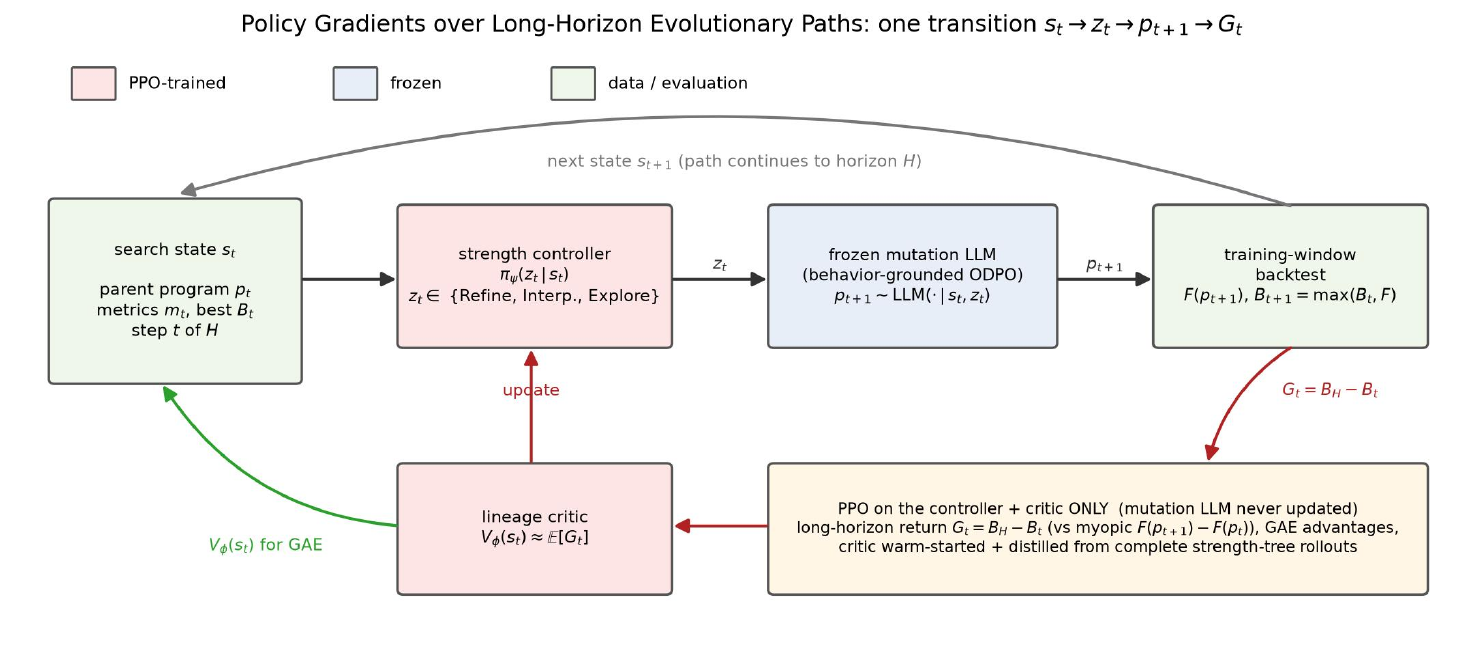}
    \caption{Lineage-Value Policy Gradients. A trainable actor LoRA and three-way head select a behaviorally calibrated mutation radius from the current program and search state. The frozen ELM mutates and compiles the child, while PPO updates only the actor LoRA and head plus a separate critic LoRA and scalar head. Path return credits downstream best-so-far progress, while depth-five trees initialize and refine the critic and supervise the strongest first branch.}
    \label{fig:method-overview}
\end{figure*}

\subsection{Path Value}
Let $\widetilde B_t$ be the best standardized fitness observed through step $t$. The best-so-far step reward is
\begin{equation}
 r_t=\widetilde B_{t+1}-\widetilde B_t.
 \label{eq:step-reward}
\end{equation}
With $\gamma=1$, the return telescopes
\begin{equation}
 G_t^{\mathrm{path}}=\sum_{k=t}^{H-1}r_k=\widetilde B_H-\widetilde B_t.
 \label{eq:path-return}
\end{equation}
An action can therefore receive positive credit after an immediate decline if a later descendant establishes a new path best. Invalid children receive an additional penalty of $0.25$ standardized units. PPO-Immediate observes the same paths but uses
\begin{equation}
 r_t^{\mathrm{imm}}=\operatorname{clip}\!\left(\widetilde F(p_{t+1})-\widetilde F(p_t),-3,3\right),
 \qquad \gamma=0.
 \label{eq:immediate-return}
\end{equation}
Equation~\ref{eq:tve} is the conceptual difference between these action values. PPO-Path directly optimizes the finite-horizon return in Equation~\ref{eq:path-return}; it does not separately estimate the residual.

\subsection{PPO-Path Algorithm}
\noindent\begin{minipage}{\columnwidth}
\hrule\smallskip
\textbf{Algorithm 1} PPO-Path for one asset-fold-seed run\par
\smallskip
\footnotesize
\begin{algorithmic}[1]

\Require frozen ELM $M_\theta$; actor LoRA/head $\pi_\psi$; critic LoRA/head $L_\phi$; $H=8$
\State initialize the archive with the pinned policy
\For{$r=1,\ldots,4$ rounds}
    \State collect eight paths in parallel
    \For{each path and $t=0,\ldots,H-1$}
        \State render $s_t$ and sample $z_t\sim\pi_\psi(\cdot\mid s_t)$
        \State generate, compile, evaluate, and record $p_{t+1}$
    \EndFor
    \State apply the fixed archive rule to 64 children
    \State expand depth-five trees from two states
    \State compute GAE and update actor and critic
\EndFor
\State select on validation, tune numeric leaves, test once
\end{algorithmic}
\smallskip\hrule
\end{minipage}

\section{Architecture and Training}
\subsection{Evolutionary Language Model}
The search representation is a natural-language trading policy. The execution representation is Genetic Programming Trading Language, a typed tree language producing Boolean long-entry, long-exit, short-entry, and short-exit signals. Compilation uses greedy decoding followed by at most two retries at temperature $0.3$. A program is invalid after three failed compilations or fewer than ten trades.

The \emph{Evolutionary Language Model} (ELM) is an offline-trained Qwen3-8B multitask model \citep{yang2025qwen3}. Its 44,736 examples cover strength-conditioned natural-language mutation, language-to-GPTL compilation, and GPTL-to-language translation. Training examples come from reversed degradation walks at offsets of one, two, and four steps. Parent and child programs are executed on a training-only probe panel, and equal-frequency tertiles of action-sequence displacement define low, medium, and high targets.

Supervised multitask learning establishes mutation and compilation competence but does not guarantee separated radii. We therefore calibrate ELM with Offset Direct Preference Optimization (ODPO) \citep{rafailov2023dpo,amini2024odpo}. ODPO uses 14,735 common-parent preference pairs, 775 validation pairs, $\beta=0.3$, margin scale $0.2$, and one epoch. The low and medium boundaries are $0.0247$ and $0.1594$. The resulting Qwen3-8B ELM is frozen during PPO in every search condition. Across 536 held-out mutations from 24 parents, median executed disagreement is $0.095$, $0.194$, and $0.326$ for Refine, Interpolate, and Explore. Parent medians have strict order in $91.7\%$ of cases, and validity exceeds $90\%$ at every strength.

\subsection{Meta-Controller}
The actor and critic are separate LoRA adapters over ELM's frozen 36-layer backbone \citep{hu2022lora}. Each uses $r=150$, $\alpha=300$ ($\alpha/r=2$), dropout $0.05$, and all seven projections in every layer: \texttt{q/k/v/o\_proj} and \texttt{gate/up/down\_proj}. Each adapter has approximately 409.2M trainable parameters, about $5\%$ of the 8.19B-parameter base. The actor maps the final state-prompt token through a zero-initialized $4096\!\rightarrow\!3$ head; the critic maps its transition-prompt token through a separate $4096\!\rightarrow\!1$ scalar head.

PPO and tree distillation update the actor LoRA and head, while Huber loss updates the critic LoRA and head. Online PPO uses clip ratio $0.2$, generalized advantage estimation with $\lambda=0.95$, entropy coefficient $0.01$, eight actor epochs, and ten critic epochs per round \citep{schulman2017ppo,schulman2016gae}.

\subsection{Critic Bootstrapping}
For each root program, we force all three first actions and expand every resulting node under all three strengths for four additional levels. A complete depth-five ternary tree has 243 leaves and requires 363 child executions before memoization. For first-action child $C_z$, the target is
\begin{equation}
 y(P,z,C_z)=\operatorname{clip}\!\left(\max_{u\in\mathcal{T}_5(C_z)}F(u)-F(C_z),-3,3\right).
 \label{eq:tree-target}
\end{equation}
The target is exact for the evaluated deterministic tree and optimistic with respect to the full stochastic ELM transition kernel. Offline critic training uses 2,109 transitions, complete-run train-validation splits, Huber loss with $\delta=1$, and three epochs. The warm start reaches held-out $R^2=0.374$ and mean absolute error $0.564$ Sharpe units. During each PPO round, two collected states receive new trees. Their targets update the critic and weight cross-entropy supervision toward the highest-valued first action.

\begin{figure*}[t]
    \centering
    \includegraphics[width=0.78\textwidth]{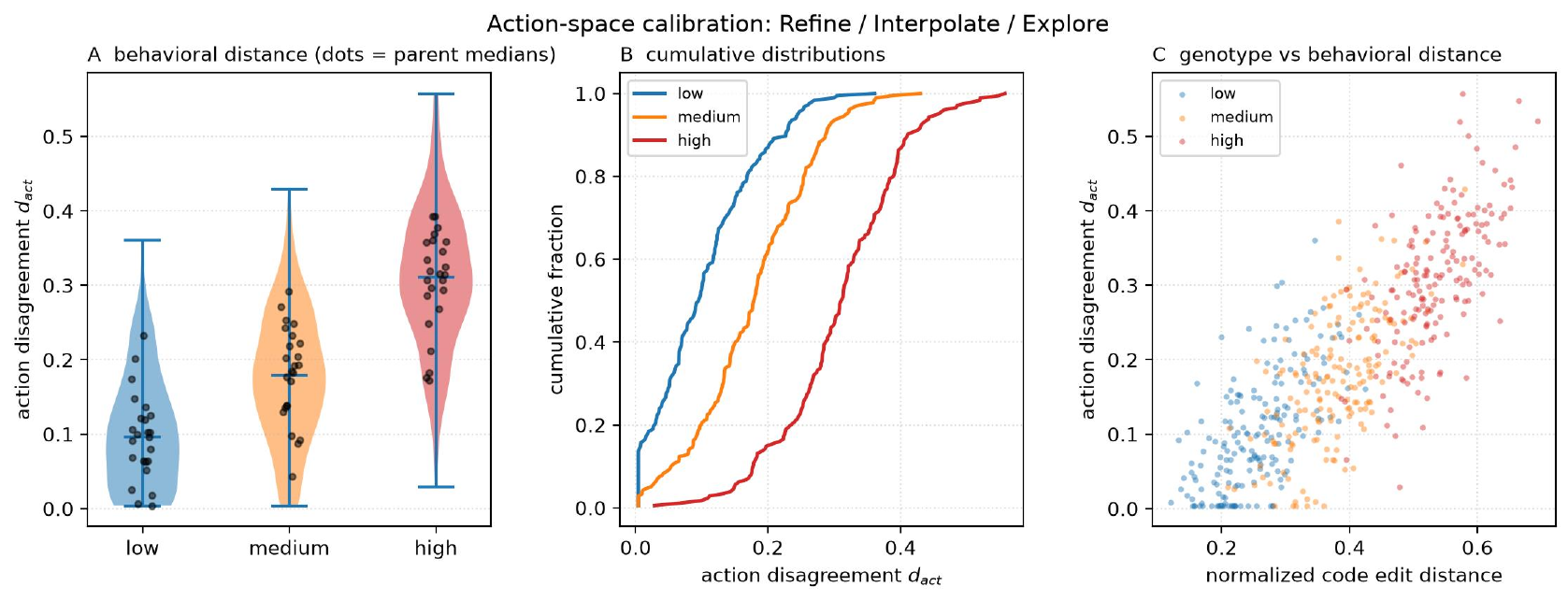}
    \caption{Behavioral calibration of the \textsc{Refine}, \textsc{Interpolate}, and \textsc{Explore} action space. (A) Realized parent--child action disagreement, with black points showing parent-level medians. (B) Empirical cumulative distributions under each requested strength. (C) Normalized code-edit distance against executed behavioral distance. The ordered distributions verify that the behaviorally calibrated ELM exposes empirically distinct semantic radii rather than nominal prompt labels.}
    \label{fig:method-diagnostics}
\end{figure*}

\section{Experimental Protocol}
\subsection{Data and Statistical Units}
We evaluate hourly continuous futures for the S\&P 500 E-mini, Silver, and 30-Year Treasury. Their histories contain 23,915, 24,020, and 23,947 bars and end on October 20, 2025. Each market has ten end-anchored rolling-origin folds. A fold contains four months of training, one month of validation, a three-day embargo, and one month of sealed test data. The ten test windows do not overlap. Training feedback drives every evolutionary, actor, and critic update. Validation selects one policy after search and tunes only numeric leaves through at most 64 deterministic backtests. Program logic remains fixed. The selected champion is evaluated once on test.

The sampling hierarchy is
\begin{equation}
 \text{asset}\rightarrow\text{fold}\rightarrow\text{seed/run}\rightarrow\text{path}\rightarrow\text{transition}.
 \label{eq:hierarchy}
\end{equation}
Transitions are never treated as independent observations. The primary unit is the complete paired asset-fold-seed run. Seven primary conditions use three assets, ten folds, and seeds $\{4,5,7\}$, giving 90 runs per condition. Three mechanism ablations use the 30 prespecified Treasury blocks. The complete matrix contains 720 runs and 184,320 archive-facing children.

Primary inference uses five planned paired contrasts of PPO-Path against PPO-Immediate, one-step control, Schedule, Uniform, and the strongest fixed radius. Assets are fixed strata. We resample complete paired blocks within strata for 10,000 hierarchical-bootstrap replicates and report mean differences, 95\% intervals, probability of superiority, and Holm-adjusted $p$-values \citep{efron1994bootstrap,holm1979sequential}. A complete-block Friedman test with paired Wilcoxon follow-ups is a secondary sensitivity analysis \citep{friedman1937rank,wilcoxon1945individual,demsar2006statistical}. Transition analyses resample complete paths within runs.

\subsection{Fitness Normalization}
Training Sharpe varies across assets and folds. Before search, each task receives a frozen training-only reference set of 64 policies. Every condition uses the same file. Fitness is standardized as
\begin{equation}
 \widetilde F_{a,f}(p)=\frac{F_{a,f}(p)-m_{a,f}}
 {\max(1.4826\,\mathrm{MAD}_{a,f},0.1)}.
 \label{eq:normalization}
\end{equation}
Rewards and critic values use this scale. Validation and test results remain in raw Sharpe units.

\subsection{Matched Baseline Matrix}
\begin{table*}[t]
\centering
\caption{Controller and baseline matrix. Every run uses 256 archive-facing child evaluations. PPO-Path and PPO-Immediate also share the same online tree budget. The one-step, four-step, and time-only ablations use the 30 Treasury blocks.}
\label{tab:matrix}
\small
\setlength{\tabcolsep}{5.0pt}
\begin{tabular}{@{}llllr@{}}
\toprule
\textbf{Condition} & \textbf{Action rule} & \textbf{Credit} & \textbf{Purpose} & \textbf{Runs}\\
\midrule
Fixed L/M/H & one constant radius & none & global-radius control & $3\times90$\\
Uniform & categorical $1/3$ & none & random action control & 90\\
Schedule & Explore$\to$Interpolate$\to$Refine & time rule & hand annealing & 90\\
One-step & learned, $H=1$ & one step & no path structure & 30\\
Four-step & learned, $H=4$ & path & horizon truncation & 30\\
PPO-Immediate & learned, $H=8$ collection & $\gamma=0$ & matched myopic PPO & 90\\
Time-only & learned, masked state & path & schedule alternative & 30\\
\textbf{PPO-Path} & learned, full state & $H=8$, $\gamma=1$ & proposed method & \textbf{90}\\
\bottomrule
\end{tabular}
\end{table*}

PPO-Path and PPO-Immediate are compute matched. Fixed, Uniform, and Schedule use the same archive-facing candidate budget but do not incur the tree and PPO training costs. Experiments used four NVIDIA RTX PRO 6000 Blackwell Server Edition GPUs, an Intel Xeon 6767P CPU, 1 TB RAM, and Ubuntu 24.04.4.

\section{Results}
\subsection{Search Performance}
\begin{figure*}[t]
    \centering
    \includegraphics[width=\textwidth]{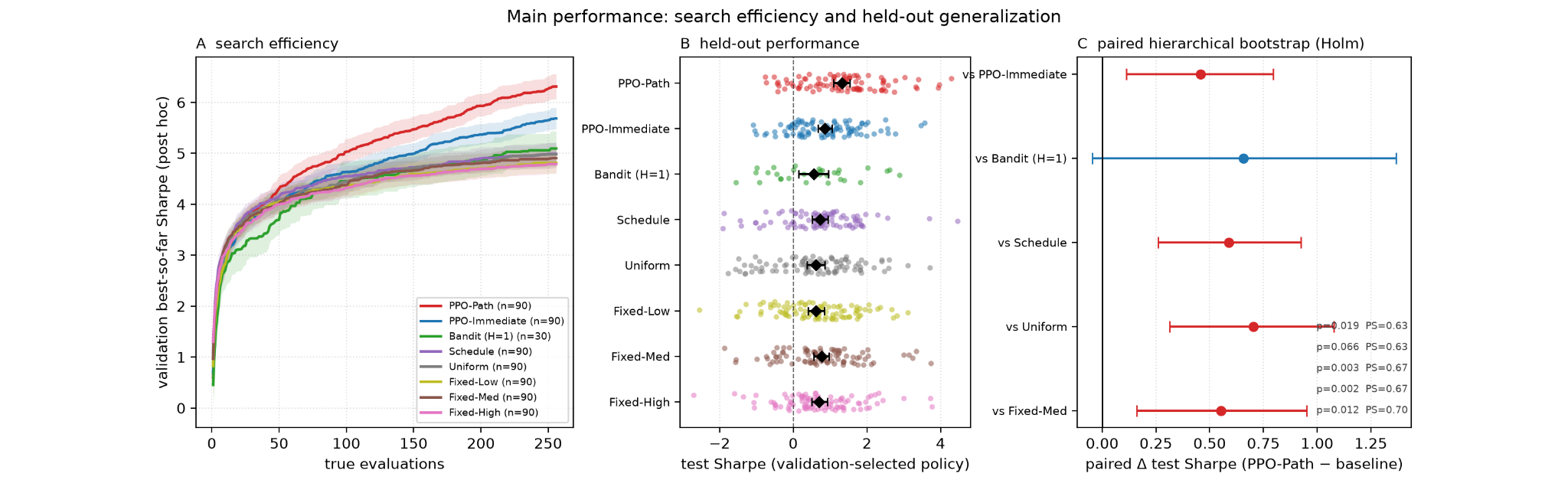}
    \caption{Primary performance. (A) Post-hoc validation best-so-far Sharpe over archive-facing children with 95\% hierarchical-bootstrap intervals. (B) Sealed-test Sharpe for each validation-selected and numerically tuned champion. Black diamonds summarize complete-run outcomes. (C) Paired PPO-Path test differences with 95\% intervals, Holm-adjusted $p$-values, and probability of superiority.}
    \label{fig:main-performance}
\end{figure*}

\begin{table}[t]
\centering
\caption{Primary paired results. Effects are PPO-Path minus baseline. Brackets give 95\% hierarchical-bootstrap CIs; $p_{\mathrm H}$ is Holm-adjusted across five planned contrasts and PS is probability of superiority.}
\label{tab:main-results}
\footnotesize
\renewcommand{\arraystretch}{1.00}
\setlength{\tabcolsep}{1.4pt}
\begin{tabular}{@{}lcc@{}}
\toprule
\textbf{Baseline} & \textbf{$\Delta$ validation AUC} & \textbf{$\Delta$ test Sharpe} \\
\midrule
\shortstack[l]{PPO-Immediate\\$n=90$} & \shortstack{$0.394\ [0.261,0.526]$\\$p_{\mathrm H}<.001$} & \shortstack{$0.459\ [0.113,0.795]$\\$p_{\mathrm H}=.0188,\ \mathrm{PS}=.633$} \\
\shortstack[l]{One-step ($H=1$)\\$n=30$} & \shortstack{$0.599\ [0.411,0.795]$\\$p_{\mathrm H}<.001$} & \shortstack{$0.657\ [-0.045,1.367]$\\$p_{\mathrm H}=.066,\ \mathrm{PS}=.633$} \\
\shortstack[l]{Schedule\\$n=90$} & \shortstack{$0.597\ [0.433,0.758]$\\$p_{\mathrm H}<.001$} & \shortstack{$0.590\ [0.261,0.924]$\\$p_{\mathrm H}=.003,\ \mathrm{PS}=.667$} \\
\shortstack[l]{Uniform\\$n=90$} & \shortstack{$0.654\ [0.486,0.807]$\\$p_{\mathrm H}<.001$} & \shortstack{$0.702\ [0.313,1.078]$\\$p_{\mathrm H}=.002,\ \mathrm{PS}=.667$} \\
\shortstack[l]{Fixed-Medium\\$n=90$} & \shortstack{$0.685\ [0.551,0.825]$\\$p_{\mathrm H}<.001$} & \shortstack{$0.553\ [0.161,0.952]$\\$p_{\mathrm H}=.012,\ \mathrm{PS}=.700$} \\
\bottomrule
\end{tabular}
\end{table}
Table~\ref{tab:main-results} reports the five planned complete-run contrasts. Figure~\ref{fig:main-performance}A shows post-hoc validation best-so-far Sharpe over the 256-child budget. Validation is computed after search and never returned to the controller. PPO-Path separates early and finishes with the highest aggregate curve. Relative to matched PPO-Immediate, validation AUC rises by $0.394$ $[0.261,0.526]$ Sharpe units with probability of superiority $0.822$. All five planned AUC contrasts remain significant after Holm correction with adjusted $p<0.001$.

The main inference comes from prespecified paired contrasts at the complete-run level. The conservative Friedman analysis of test Sharpe on the 30 blocks shared by every method gives $p=0.265$. This sensitivity test removes 60 non-Treasury blocks and evaluates a broader omnibus hypothesis, so it does not replace the paired three-market comparison.

\subsection{Horizon and State Dependence}
\begin{figure*}[t]
    \centering
    \includegraphics[width=\textwidth]{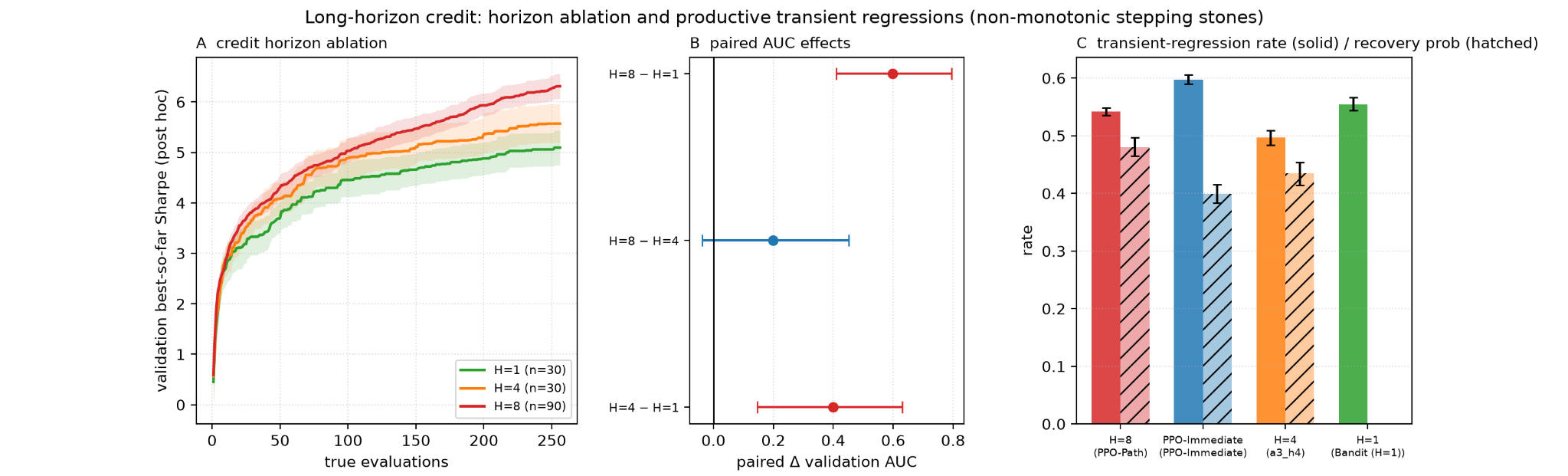}
    \caption{Credit horizon and recoverability. (A) Validation best-so-far curves for one-step, four-step, and eight-step credit under the same archive-facing budget. (B) Paired AUC differences with 95\% hierarchical-bootstrap intervals. (C) Temporary-regression rate and conditional recovery probability. Recovery requires a later descendant to exceed the path best observed before the regression.}
    \label{fig:horizon-credit}
\end{figure*}

Panels A and B of Figure~\ref{fig:horizon-credit} locate the temporal effect. Moving from one to four steps raises validation AUC by $0.399$ $[0.146,0.630]$, adjusted $p=0.0036$. Moving from one to eight steps raises it by $0.599$ $[0.411,0.795]$, adjusted $p<0.001$. The additional eight-step gain over four steps is $0.200$ $[-0.038,0.451]$ with $p=0.0954$. The strongest conclusion is therefore one-step versus multistep credit. The longest tested horizon is descriptively best and is not established as universally optimal.

A temporary regression satisfies $F(p_{t+1})<B_t$ and recovers when a later descendant exceeds $B_t$. PPO-Path produces fewer regressions than PPO-Immediate, $54.2\%$ $[53.5\%,54.8\%]$ versus $59.8\%$ $[58.9\%,60.6\%]$, and recovers from more of them, $48.0\%$ $[46.5\%,49.7\%]$ versus $39.9\%$ $[38.4\%,41.5\%]$. Its regressions are deeper, $0.94$ versus $0.77$ Sharpe units, but they recover sooner, $1.93$ versus $2.20$ later steps, and yield larger post-recovery improvement, $0.535$ versus $0.251$. Long-horizon control therefore makes non-monotonic search more selective.

The actor also reads more than the clock. A multinomial action model using remaining budget, parent-fitness percentile, stagnation, recent improvement, and prespecified interactions fits sampled actions better than a time-only model. The likelihood-ratio statistic is $544.74$ with $p<10^{-100}$, and grouped held-out log loss improves from $0.7465$ to $0.7350$. Leave-one-asset-out fits preserve the difference. The actor follows a broad exploration-to-refinement tempo, while program quality and stagnation shift the action mixture within the same remaining budget.

\subsection{Interventional Action Values}
\begin{figure*}[!t]
    \centering
    \includegraphics[width=0.95\textwidth]{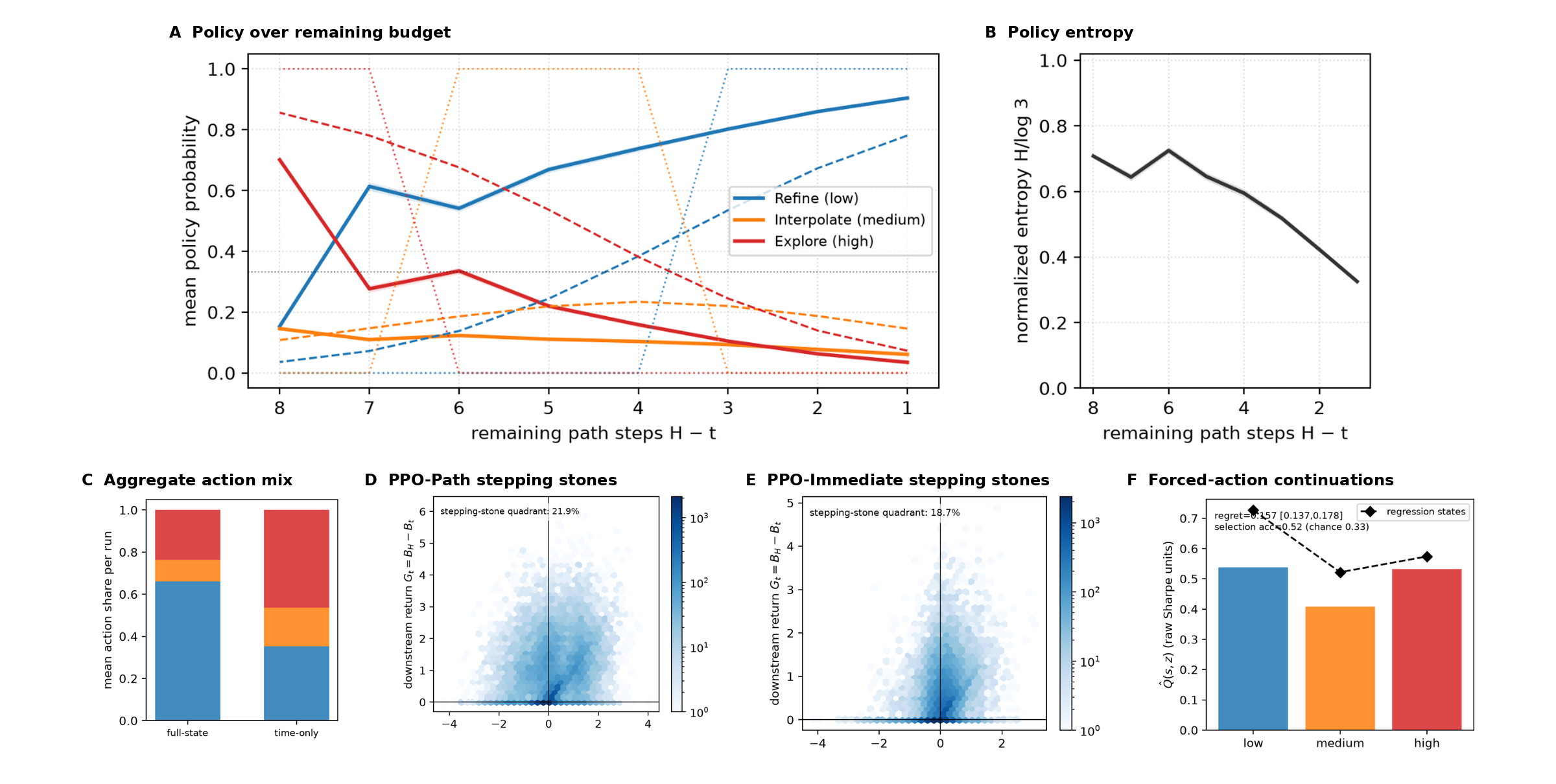}
    \caption{Structural mechanism diagnostics. (A--C) Budget- and state-dependent control. (D--E) Immediate change versus downstream return. (F) Matched forced-action continuation values. Upper-left points are temporary regressions followed by positive downstream return.}
    \label{fig:mechanism-diagnostics}
\end{figure*}

For 720 held-out states, we force each first action and run four matched stochastic continuations using the same controller and paired random seeds. Figure~\ref{fig:mechanism-diagnostics}F reports these interventional finite-horizon values. The controller selects the highest estimated action in $51.9\%$ $[48.2\%,55.8\%]$ of states, above the $33.3\%$ chance rate. Mean regret is $0.157$ $[0.137,0.178]$ Sharpe units. Across all states, Refine and Explore have nearly equal mean continuation value, $0.538$ and $0.533$, while Interpolate obtains $0.408$. Among states already in regression, Refine is strongest at $0.727$. The learned rule cannot be reduced to choosing the largest available mutation.

\subsection{Held-Out Critic Validation}
On 23,040 states excluded from critic fitting and PPO updates, critic predictions have mean run-level Spearman correlation $0.597$ $[0.585,0.608]$ with realized path returns, explained variance $19.3\%$, and mean absolute error $0.509$ Sharpe units. Calibration quintiles are monotonically ordered. The highest quintile overpredicts realized return, so the critic is an informative ranker rather than an exact oracle. After controlling for parent fitness, path best, remaining budget, stagnation, and recent improvement, the critic retains coefficient $0.361$ $[0.332,0.388]$ and incremental partial $R^2=0.116$. Full code context therefore contributes future-yield information beyond simple state proxies.

\subsection{Financial Generalization}
Figure~\ref{fig:main-performance}B and Table~\ref{tab:main-results} report the one sealed-test champion per run. Mean Sharpe rises from $0.862$ under PPO-Immediate to $1.321$ under PPO-Path. The paired gain is $0.459$ $[0.113,0.795]$, with probability of superiority $0.633$ and Holm-adjusted $p=0.0188$. Gains over Schedule, Uniform, and Fixed-Medium also remain significant; the one-step contrast is positive but uncertain.

Secondary means also favor PPO-Path. Annualized return is $22.8\%$ versus $15.3\%$, Sortino is $2.462$ versus $1.623$, maximum drawdown is $3.87\%$ versus $4.29\%$, and positive test Sharpe occurs in $86.7\%$ versus $81.1\%$ of runs. These endpoints are descriptive. Within each asset-fold-seed block, paired daily returns are additionally evaluated with a Ledoit-Wolf studentized circular-block bootstrap using automatic block length and 2,000 replicates \citep{ledoit2008sharpe}. Fold return streams are not concatenated. The evidence supports aggregate policy-discovery gains across rolling origins and does not imply superiority in every one-month test period.

\section{Discussion and Limitations}
LVPG treats mutation value as the combination of immediate fitness and finite-budget continuation value. In the matched PPO comparison, extending the return horizon improves search AUC, held-out Sharpe, and recovery from temporary regressions without changing the mutation operator or lineage supervision. The learned controller follows a broad exploration-to-refinement rhythm while adjusting to program quality and stagnation, and no mutation radius is uniformly best.

\subsection{Limitations}
ELM fixes the content distribution and the three calibrated mutation radii. This isolates temporal control but prevents online improvement of edits and leaves a coarse, domain-specific action space. LVPG also combines path PPO, critic learning, and first-action distillation. Matching critic and tree supervision between PPO-Path and PPO-Immediate isolates return horizon, not the independent necessity of each component. The critic learns optimistic maxima from deterministic depth-five trees rather than expected stochastic returns. Its held-out ranking signal is useful but imperfectly calibrated. PPO-Path and PPO-Immediate are compute matched; fixed, uniform, and scheduled arms are matched only in archive-facing children. Finally, the study uses three futures markets, one policy language, horizons through eight steps, and one-month test windows. Fixed costs omit market impact, capacity, latency, and order-book effects. The results establish finite-budget search gains under controlled historical evaluation, not live profitability or global optimality. Within these limits, the experiments show that credit over reachable descendants can improve finite-budget evolutionary search beyond credit assigned only to the immediate child.

\section{Conclusion}
In this work, we identify a central limitation of immediate-return mutation control: a mutation can be a weak child yet a valuable ancestor. When credit is assigned only from offspring fitness, actions that create recoverable temporary regressions are treated as destructive even when they open productive future lineages. Framing finite-budget search as an investment problem, we formalize this delayed utility as the time value of evolution within a finite-horizon Markov decision process. We introduce Lineage-Value Policy Gradients (LVPG), an actor-critic framework that values mutation decisions through the finite-budget futures they make reachable.

LVPG pairs an offline-trained Qwen3-8B mutation operator frozen during PPO with separate LoRA actor and critic adapters. The actor modulates behaviorally calibrated mutation intensity, while the critic is bootstrapped from multistep mutation trees to estimate finite-horizon lineage potential. Across 90 paired runs under matched operators, lineage supervision, folds, seeds, and budgets, path-based credit increases validation best-so-far AUC by 0.394 Sharpe units and raises mean sealed-test Sharpe from 0.862 to 1.321. LVPG also produces fewer temporary regressions and recovers from them more often. Long-horizon credit therefore improves the selectivity of non-monotonic search rather than simply encouraging larger or more frequent exploratory moves. These findings support a broader principle for finite-budget optimization. Immediate offspring fitness is an incomplete measure of an action’s value because it omits the future search opportunities that action creates. LVPG provides a general blueprint for generative program search: preserve a capable variation engine, learn state- and budget-dependent control over its mutation scale, and assign credit over the lineages produced by those decisions. The present evidence establishes stronger policy discovery within matched finite budgets and motivates evaluation across other program-search domains, longer horizons, and richer action spaces.

\clearpage
\bibliography{references}
\end{document}